\documentclass[letterpaper, 10 pt, conference]{ieeeconf}  

\IEEEoverridecommandlockouts                              

\usepackage{booktabs}                                   
\usepackage{multirow}                                   
\usepackage{makecell}                                   
\usepackage{tablefootnote}                              
\usepackage[symbol]{footmisc}                           
\usepackage{amsmath,amssymb}                                    
\usepackage{xcolor}                                     
\let\labelindent\relax              
\usepackage{enumitem}                                   
\usepackage{subcaption}                                 
\usepackage[misc]{ifsym}     

\usepackage{hyperref}   
\usepackage{graphicx}
\usepackage{changes}

\usepackage[misc]{ifsym}

\usepackage{hyperref}

\usepackage{color}

\newcommand{\para}[1]{\subsubsection{#1}}

\newcommand{\eat}[1]{}                                  

\newcommand{\ummm}{SM$^3$} 
\newcommand{\mma}{MMArt}

\title{\LARGE \bf
  {\ummm}: Self-Supervised Multi-task Modeling with Multi-view 2D Images for Articulated Objects
}

\author{Haowen Wang$^{1}$, Zhen Zhao$^{2}$, Zhao Jin$^{2}$, Zhengping Che$^{2}$, Liang Qiao$^{1}$, \\
Yakun Huang$^{1}$, Zhipeng Fan$^{1}$, Xiuquan Qiao$^{1, \dagger}$, and Jian Tang$^{2, \dagger}$
\thanks{$^{1}$State Key Laboratory of Networking and Switching Technology, Beijing University of Posts and Telecommunications, China
        {\tt\small \{hw.wang, LiangQ, ykhuang, fzp, qiaoxq\}@bupt.edu.cn}}
\thanks{$^{2}$Midea Group, China
        {\tt\small \{zhaozhen8, jinzhao1, chezp, tangjian22\}@midea.com}}
\thanks{
    Work done during Haowen Wang's internship at Midea Group.
}
\thanks{
    $^\dagger$Corresponding authors: Xiuquan Qiao and Jian Tang.
}
}

\begin{document}

\maketitle
\thispagestyle{empty}
\pagestyle{empty}

\begin{abstract}
  Reconstructing real-world objects and estimating their movable joint structures are pivotal technologies within the field of robotics. 
Previous research has predominantly focused on supervised approaches, relying on extensively annotated datasets to model articulated objects within limited categories. 
However, this approach falls short of effectively addressing the diversity present in the real world. 
To tackle this issue, we propose a self-supervised interaction perception method, referred to as {\ummm}, which leverages multi-view RGB images captured before and after interaction to model articulated objects, identify the movable parts, and infer the parameters of their rotating joints.
 By constructing 3D geometries and textures from the captured 2D images, {\ummm} achieves integrated optimization of movable part and joint parameters during the reconstruction process, obviating the need for annotations. 
 Furthermore, we introduce the {\mma} dataset, an extension of PartNet-Mobility, encompassing multi-view and multi-modal data of articulated objects spanning diverse categories. 
 Evaluations demonstrate that {\ummm} surpasses existing benchmarks across various categories and objects, while its adaptability in real-world scenarios has been thoroughly validated.
\end{abstract}

\section{Introduction}
Recreating real-world objects in a virtual environment and predicting how the open parts of an item will move is a critical step in helping robots complete daily household tasks~\cite{liu2022akb,mittal2022articulated,bao2023dexart,schiavi2023learning}. 
The design process of an object often requires designers to create its geometric shape and motion structure of movable parts from scratch. 
The same is true for robot learning systems.
 By using methods such as implicit neural representations~(INR)~\cite{park2019deepsdf, chen2021learning, Chan_2022_CVPR} and neural radiance fields~(NeRF)~\cite{mildenhall2021nerf, muller2022instant, xu2022point}, it has been possible to achieve a relatively complete reconstruction of static objects in the environment.
 However, accurately modeling articulated objects with movable components remains a challenging problem in this field. 
 This necessitates the accurate identification of attributes such as joint position, directions, and active component areas.
 
Recent studies~\cite{wang2019shape2motion, yan2019rpm} adopted PointNet-like structure~\cite{qi2017pointnet} to predict movable parts and their associated motion parameters directly from the point cloud of articulated objects. However, these methods usually highly rely on the complete point clouds of objects, resulting in higher requirements on equipment and difficult implementation in the real world.
To address this issue, some techniques~\cite{li2020category, heppert2023carto} have embraced the Normalized Object Coordinate Space (NOCS) approaches~\cite{wang2019normalized, wang2023dtf}  to canonicalize object scales and joints,
enabling the reconstruction of articulated objects or joint parameter estimation from a single-view image.
These methods typically underperform with novel shapes and exhibit certain limitations.

The recently introduced Ditto~\cite{jiang2022ditto} offers a fresh perspective, demanding point clouds before and after object interaction to decode its motion and geometric characteristics. 
It employs implicit neural representations for 3D object reconstruction and integrates part segmentation and joint parameters through joint optimization. 
However, it still struggles to model entirely new classes of objects without training and to accurately estimate the associated motion states.
Furthermore, these methods rely on high-cost, large-scale datasets of articulated objects, which necessitate precise 3D part models and joint annotations.

\begin{figure}[t]
  \centering
  \includegraphics[width=1\linewidth]{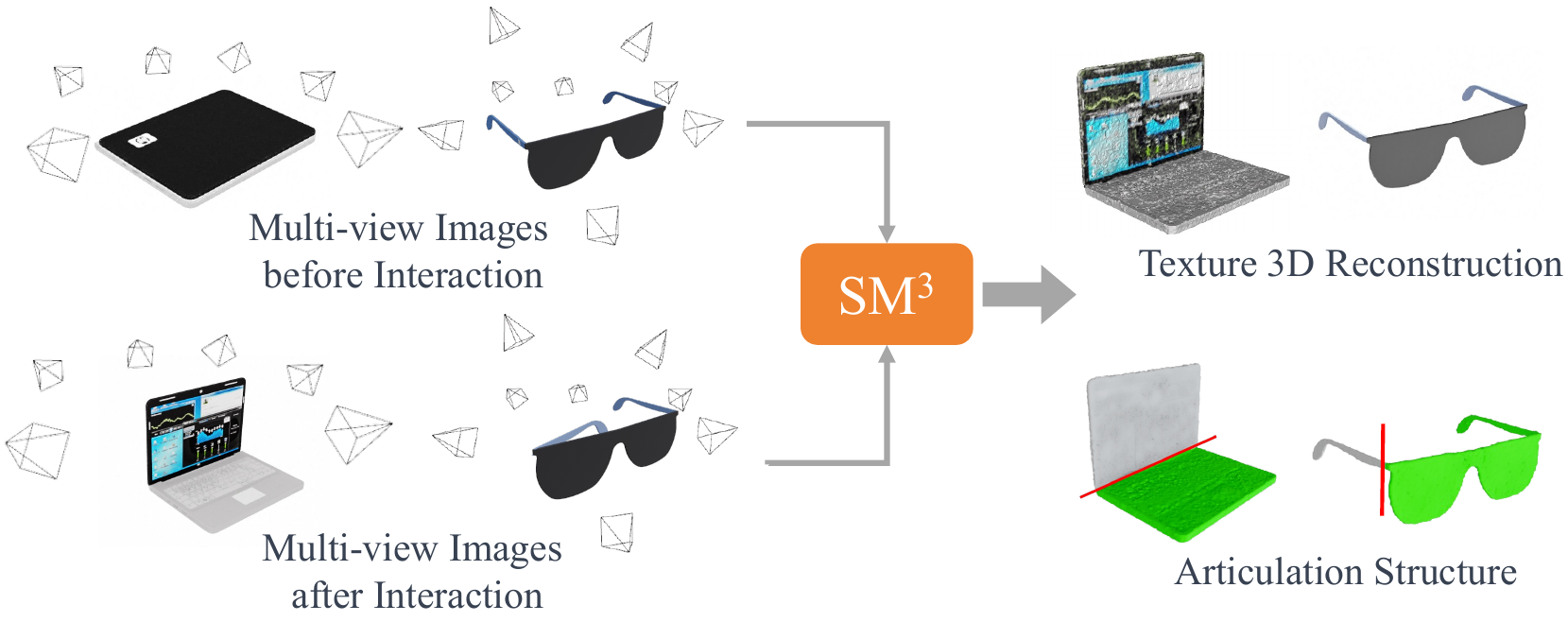}
  \caption{Our proposed {\ummm} enables textured 3D reconstruction and articulation structure estimation solely from multi-view images captured before and after object interaction.
  }
  \label{fig:first}
\end{figure}
 
In parallel, embodied AI~\cite{ahn2022can, driess2023palm, ha2023scaling, brohan2023rt} has showcased a remarkable trajectory. Yet, this trajectory necessitates extensive data resources, particularly in the domain of 3D scene acquisition, encompassing articulated objects. Consequently, the need arises for a method that can effectively reconstruct data for the various objects present in a scene, thereby facilitating further advancements in this field.

In light of the aforementioned challenges, 
we propose a novel approach called Self-Supervised Multi-task Modeling with Multi-view 2D Images for Articulated Objects~({\ummm}).
Utilizing multi-view RGB images of articulated objects taken both pre- and post-interaction, {\ummm} achieves textured 3D reconstruction, segmentation of the movable part, and the estimation of corresponding rotational joint parameters, as shown in Fig.~\ref{fig:first}.
Our 3D reconstruction methodology builds upon the Nvdiffrec~\cite{munkberg2022extracting} framework, employing a deformable tetrahedral grid and computing post-rendering image losses for objects before and after interaction. 
This tetrahedral structure serves as the foundation for subsequent movable part segmentation and joint parameter optimization.
To generate reliable constraints for integrated optimization and mitigate convergence issues, we design two algorithmic workflows that analyze geometric structure differences between objects in their pre- and post-interaction states.
These workflows facilitate the generation of movable part segmentation priors and rotational joint position and direction candidates.
Furthermore, we introduce the patch-split method to refine the original image loss, thereby enhancing segmentation accuracy. 
After several steps in the process, we render the pre-interaction tetrahedral grid of the movable parts as they rotate through the joints, converting them to RGB images. 
To finalize the optimization of the accuracy of movable part segmentation and rotational joint parameters, we calculate the loss against post-interaction object-sampled RGB images.
Given the scarcity of multi-view RGB image datasets specifically designed for articulated objects, 
we have meticulously curated a pioneering dataset named {\mma}, 
building upon the PartNet mobility dataset. 
This multi-view, multi-modal, and multi-state dataset encompasses a diverse array of articulated objects spanning multiple categories.
 It caters to our evaluation requirements and readily adapts to facilitate the training and testing of other methods within this domain.

Our contributions can be summarized as follows:

\begin{itemize} [itemsep=1pt,topsep=1pt,parsep=1pt,leftmargin=10pt]
    
    \item We introduce {\ummm}, a pioneering self-supervised multi-task method that simultaneously learns textured 3D reconstruction, movable part segmentation, and rotating joint parameters for articulated objects.

    \item We devise two algorithmic workflows to effectively generate active part segmentation priors and joint candidates to further refine accuracy through optimization strategies.

    \item We present the {\mma} dataset that supports comprehensive evaluations for articulated object modeling.
    
    \item Our experimental results demonstrate the effectiveness of {\ummm} in accurately modeling articulated objects, surpassing recent SOTA methods by a significant margin.

\end{itemize}

\section{Related Work}
\para{3D Reconstruction for Multi-View Images} Pioneer methods~\cite{agarwal2011building,schonberger2016pixelwise} perform geometric reconstruction through stereo matching of multi-view RGB images. However, they rely heavily on extensive training data and tend to produce gaps or holes in areas with weak textures.
Modern approaches~\cite{oechsle2021unisurf,wang2021neus} pivot to implicit representations. 
NeRF~\cite{mildenhall2021nerf}, for instance, exploits radiance fields for unseen view synthesis but doesn't yield directly usable textured 3D models. 
While UNISURF~\cite{oechsle2021unisurf} and NeuS~\cite{wang2021neus} refine reconstructions, they cater only to static objects. 
A-SDF~\cite{mu2021sdf}, though innovative in decoupling shape and joint features, remains untextured and less detailed. 
Contrarily, capitalizing on the Nvdiffrec~\cite{munkberg2022extracting} paradigm, our method adeptly refines 3D geometries and textures, concurrently optimizing articulation structures.

\para{Motion Structure Estimation}  
Methods such as~\cite{abbatematteo2019learning,jain2021screwnet,jain2022distributional} focus on joint estimation and component segmentation for objects. 
However, they heavily depend on the input of complete point clouds, often neglecting object reconstruction.
ANCSH~\cite{li2020category} and OMAD~\cite{xue2021omad} predict segmentation and joint parameters from single-view point cloud but struggle with untrained objects. 
Ditto~\cite{jiang2022ditto}integrates joint prediction with 3D reconstruction but demands multi-state joint point clouds during training. 
These methods require extensive annotated data and precise 3D models, which are challenging to obtain in real-world scenarios.
In contrast, our approach, utilizing multi-view RGB images, seamlessly integrates textured object reconstruction, component segmentation, and joint prediction.

\para{Datasets for Articulated Objects.}  
Currently, several datasets offer 3D models of articulated objects.
RPM-Net~\cite{yan2020rpm} offers 949 joint objects across 43 categories. RBO~\cite{martin2019rbo} provides RGB-D videos of 14 articulated objects, while Shape2Motion~\cite{wang2019shape2motion} boasts 2,440 joint objects in 45 categories. 
PartNet~\cite{mo2019partnet} emphasizes semantic segmentation, whereas PartNet-Mobility~\cite{xiang2020sapien} enriches PartNet and ShapeNet~\cite{chang2015shapenet} with 2,346 joints across 46 categories. 
Moreover, these datasets lack multi-view and multi-modal data. 
To address this gap, we introduce a new dataset to enrich training and testing tasks in this domain.

\section{Methodology}
\begin{figure*}[t]
  \centering
  \includegraphics[width=1\linewidth]{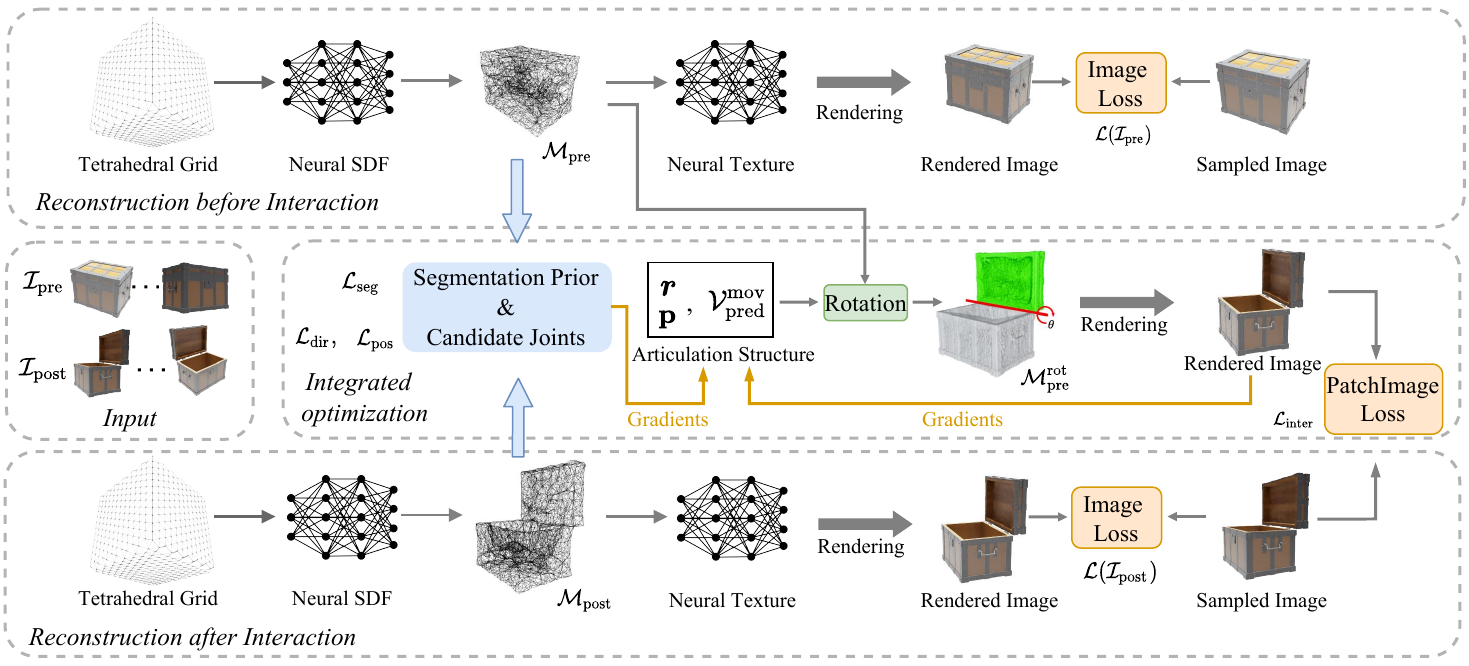}
  \caption{Architecture Overview of proposed {\ummm}.
  }
  \label{fig:architecture}
\end{figure*}


 \subsection{Overview}
The goal of our work is to actively construct textured virtual geometric models of real-world articulated objects and estimate the articulation structure. 
An overview of our framework is shown in Fig.~\ref{fig:architecture}. 
For an articulated object in the scene, we initially capture $n$ RGB images from varied viewpoints, denoted as $\mathcal{I}_\text{pre} = \{I_i\}_n$. 
After actively manipulating the object, leading to rotation of its movable part, we obtain another set of $m$ images, $\mathcal{I}_\text{post} = \{I_j\}_m$.
 Our method processes these two sets of observations to construct a relatively detailed 3D texture model of the articulated object, while accurately estimating the rotational joints and segmenting the movable part.

For building the geometric model, we use a method based on the deformable tetrahedral grid~\cite{gao2020learning, shen2021deep, munkberg2022extracting} to render the model $\mathcal{T} = (\mathcal{V}, \mathcal{F})$. $\mathcal{V} \in \mathbb{R}^{n \times 3}$ represents the vertex of the model, $n$ is the number of vertices.
$\mathcal{F}\in \mathbb{N}^{m \times 4}$ is represented as the tetrahedral face of the model, and each tetrahedron $\mathcal{F}$ contains four vertices. 
Movable part are modeled using a vertex subset, $\mathcal{V}_{s} \in \mathbb{R}^{n' \times 3}$.

For joint construction, 
we follow the definition method in line with ~\cite{li2020category, jiang2022ditto}, 
where kinematic constraints between object components are represented by parameterizing the joints. 
A revolute joint is characterized by its rotation axis direction $\boldsymbol{r} \in \mathbb{R}^{3}$ and a pivot point $\mathbf{p} \in \mathbb{R}^{3}$ on this axis. 
Notably, both the rotation axis and pivot point remain invariant to interactions.

\label{sec:3.1_problem-def}

\subsection{Reconstruction Based on Tetrahedron}

To enhance the prediction of object geometric reconstruction, aiding subsequent joint and part segmentation, we employ Nvdiffrec~\cite{munkberg2022extracting}. 
This approach leverages differentiable tetrahedrons to optimize the topology and texture of the initial tetrahedra directly, as shown in Fig.~\ref{fig:architecture}.

Given multi-view sampled images $\{I\}$, an implicit neural network  determines the deformation directions $v$ and the Signed Distance Function~(SDF) values $s$ for each vertex of the initial tetrahedral grid:
\begin{equation}
(v, s) = f_\phi(I),
\end{equation}
where $f_\theta$ is the multi-layer perception with parameters $\phi$.

Utilizing $v$ and $s$, the tetrahedral grid $\mathcal{M}$ refines to capture the object's geometry better. 
Concurrently, the texture image $T$ and lighting information $L$ are inferred from $I$. With these, a differentiable rasterizer~\cite{munkberg2022extracting} renders an RGB image:
\begin{equation}
 I_r^{c} = R\left(\mathcal{M}, T, L; \Phi\right), 
\end{equation}
where $ I_r^{c} $ is the rendered RGB image for a given camera pose $c$ and $\Phi$ are the parameters of the rasterizer.

The optimization aims to minimize the image loss between the rendered images ${I}_r$ and the input images ${I}_{gt}$:
\begin{equation}
\mathcal{L}_{\text{render}} = \frac{1}{n \cdot hw} \sum_{c=1}^n \sum_{i=1}^h \sum_{j=1}^w (I_{r,ij}^c - I_{gt,ij}^c),
\end{equation}
where $h$ and $w$ denote the dimensions of each input image,  $I_{r,ij}^c$ and $I_{gt,ij}^c$ are pixel values at position $(i, j)$ in the rendered and input images, respectively, for camera pose $c$.
Through iterative propagation across all perspective images $\mathcal{I} = \{I_i\}_n$, we obtain the tetrahedral grid representing the object's geometry and its associated texture.

Leveraging this reconstruction approach, we derive two 3D geometric models: $\mathcal{M}_{\text{pre}}$ from the pre-interaction images $\mathcal{I}_\text{pre}$ and $\mathcal{M}_{\text{post}}$ from the post-interaction images $\mathcal{I}_\text{post}$.
\label{sec:3.2_recon}

\subsection {Movable Part Segmentation Prior}

\begin{figure}[t]
  \centering
  \includegraphics[width=1\linewidth]{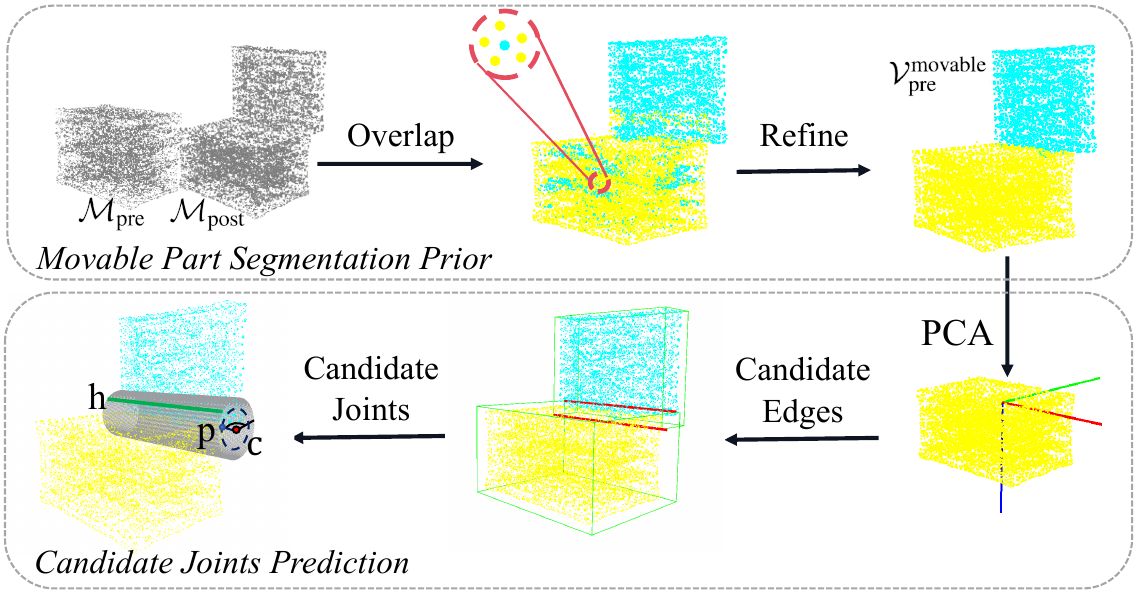}
  \caption{
Algorithmic workflow for Movable Part Segmentation Prior and Candidate Joints Prediction.
  }
  \label{fig:candidate}
\end{figure}

For articulated object modeling, it's pivotal to derive segmentation priors for the movable part. 
We ascertain the overlap between the 3D models $\mathcal{M}_{\text{pre}}$ (pre-interaction) and $\mathcal{M}_{\text{post}}$ (post-interaction) to determine segmentation prior, as shown in Fig.~\ref{fig:candidate}.

Specifically, we analyze vertices from the tetrahedral models of both states, $\mathcal{V}_{\text{pre}}$ and $\mathcal{V}_{\text{post}}$. 
For a vertex $v_i$ in $\mathcal{V}_{\text{pre}}$, it's labeled as static (0) if its distance to the nearest vertex in $\mathcal{V}_{\text{post}}$ is below a threshold $\tau$, and movable (1) otherwise. This is expressed as:
\begin{equation}
\text{label}(v_i) =
\begin{cases}
0 & \text{if } \min_{v_j \in \mathcal{V}_{\text{post}}} \text{dist}(v_i, v_j) < \tau \\
1 & \text{otherwise}
\end{cases}
\end{equation}
where we set $\tau = 0.04$ in our study.

During segmentation refinement, we rectify potential errors from overlaps using neighborhood analysis. 
For each vertex, its label is determined by the majority label of its neighboring vertices. 
If the majority of a vertex's neighbors are labeled as static, then the vertex itself is labeled as static. 
Conversely, if most neighbors are movable, the vertex is labeled as movable.

From the described workflow, we derive the segmentation prior for the movable part of the pre-interaction object's tetrahedral grid vertices, denoted as $\mathcal{V}_{\text{pre}}^{\text{mov}}$.
In the optimization phase (Section~\ref{sec:3.5_joint-optim}), we use a cross-entropy loss $L_{\text{seg}}(\mathcal{V}_{\text{pred}}^{\text{mov}},  \mathcal{V}_{\text{pre}}^{\text{mov}})$ to guide the prediction of movable part vertices.

\label{sec:3.3_priori-seg}

\subsection{Candidate Joints Prediction}

Identifying the revolute joints in real-world articulated objects is challenging. Our method, illustrated in Fig.~\ref{fig:candidate}, assumes joint directions often align with the object's primary planes~\cite{lee2020aerial}.

Applying Principal Component Analysis (PCA) to the static part's point cloud from Section~\ref{sec:3.3_priori-seg}, we establish a local coordinate system. Within this, we compute Axis-Aligned Bounding Boxes (AABBs) for both movable and static parts, favoring them over Oriented Bounding Boxes (OBBs). Candidate joint positions are identified by matching parallel edges from both parts' bounding boxes, ranked using:
\begin{equation}
\text{rank} = \text{rank}{\text{dist}} + w \times (\text{rank}{\text{e1}} + \text{rank}{\text{e2}}).
\end{equation}
Here, $\text{rank}{\text{dist}}$ is distance-based, 
while $\text{rank}{\text{e1}}$ and $\text{rank}{\text{e2}}$ consider the count of randomly sampled points along each edge that have both static and movable components in their neighborhood.

The top-ranked edge pair suggests a guiding cylinder for the joint's position and direction. Using the cylinder's normal vector $\mathbf{h}$, the joint's position is transformed into Cartesian coordinates, yielding the pivotal point $\mathbf{p}$.

The joint position and direction loss functions are:
\begin{equation}
\mathcal{L}_{\text{pos}} = \max(0, ||\mathbf{p} - \mathbf{c} ||2 - R),
\end{equation}
and
\begin{equation}
\mathcal{L}_{\text{dir}} = 1 - \frac{\boldsymbol{r} \cdot \mathbf{h}}{|\boldsymbol{r}| |\mathbf{h}|},
\end{equation}
ensuring alignment with our assumptions.
\label{sec:3.4_candi-joint}

\subsection{Integrated Optimization}

To integrate optimization of the predicted joint parameters, specifically the direction vector $ \mathbf{r} $ and the pivotal point $ \mathbf{p}$, alongside movable part segmentation, 
we employ rigid rotation transformations on the movable components.

For each vertex in the movable component of the pre-interaction tetrahedral grid, denoted as \( \mathbf{v} \in \mathcal{V}_{\text{pre}}^{\text{movable}} \), we compute its post-rotation position \( \mathbf{v}' \) using the formula:
\begin{equation}
\mathbf{v}' =  \mathbf{R}(\mathbf{r}, \theta)(\mathbf{v} - \mathbf{p}) + \mathbf{p}
\end{equation}
Here, \( \mathbf{R}(\boldsymbol{r}, \theta) \) is the rotation matrix derived from the Rodrigues formula~\cite{wandt2021canonpose}, \( \theta \) signifies the rotation angle.
Through the above operation, we transform the tetrahedra from their pre-interaction state $\mathcal{M}_{\text{pre}}$ to their post-interaction state $\mathcal{M}^\text{rot}_{\text{pre}}$.

Subsequently, analogous to Section~\ref{sec:3.2_recon}, we render the deformed tetrahedral structure into an RGB image and compute the loss concerning the captured RGB image post-interaction.
 Importantly, during the optimization process, we refrain from propagating gradients to the SDF values and offset vector predictions of the tetrahedral vertices, preserving the geometric topology of the reconstructed object.

In tandem with our approach, we segment the image into patches and introduce the PatchImage Loss.
By computing the loss for each patch and subsequently aggregating them, we derive the comprehensive image loss. 
This strategy adeptly mitigates the influence of minor noise or misalignments in the rotation axis, facilitating a granular examination of distinct image regions. 
The integrated loss function is articulated as:
\begin{equation}
\begin{aligned}
\mathcal{L}_{\text{inter}} &= \frac{1}{n \cdot hw} \sum_{c=1}^n \sum_{\text{patches}} \frac{1}{hw} \sum_{i=1}^h \sum_{j=1}^w \Delta R_{ij}^c \\
\Delta R_{ij}^c &= R(\mathcal{M}^\text{rot}_{\text{pre}}, T, L; \Theta){ij}^c - {I_{ij}^c}_{post}
\end{aligned}
\end{equation}
The overall loss function is formulated as follows:
\begin{equation}
\mathcal{L}_{\text{total}} = \lambda_1 \mathcal{L}_{\text{seg}} + \lambda_2 \mathcal{L}_{\text{pos}} + \lambda_3 \mathcal{L}_{\text{dir}} + \lambda_4 \mathcal{L}_{\text{inter}} 
\end{equation} 
where $\lambda_1$, $\lambda_2$, $\lambda_3$, and $\lambda_4$ equal 1, 2, 2, and 10, respectively, in this study.
Utilizing this loss computation, we achieve integrated optimization of movable part and joint parameters, guided by geometric priors and RGB supervision.

\label{sec:3.5_joint-optim}

\section{The MMArt Dataset}
\begin{figure*}[t]
  \centering
  \includegraphics[width=1\linewidth]{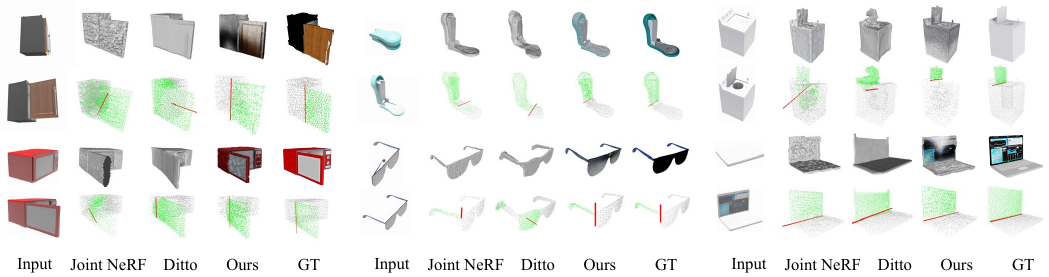}
  \caption{Visualization of 3D reconstruction and articulation. Movable parts are shown as green point clouds for clarity.}
  \label{fig:visual}
\end{figure*}
 
Several existing datasets~\cite{yan2020rpm, mo2019partnet, xiang2020sapien} provide component models and joint parameters for articulated objects. 
However, a clear void exists for datasets dedicated to multi-view 2D image-driven 3D reconstruction of these entities. 
Bridging this void, we create the Multi-Modal Articulated Objects Dataset~(\mma), 
an extension of the PartNet-Mobility~\cite{xiang2020sapien} dataset, enriched using the Isaac Sim simulation environment~\cite{isaac}.

Specially, we meticulously selected thirteen categories: box, dishwasher, door, eyeglasses, laptop, lighter, microwave, safe, stapler, storage furniture, toilet, oven, and washing machine from the PartNet-Mobility~\cite{xiang2020sapien} dataset.
Within each of these categories, we carefully chose eight unique objects for inclusion.
To ensure high data fidelity, we utilized Nvidia Isaac Sim~\cite{isaac}, a robust robotics simulation platform and synthetic data generation tool, to sample from the PartNet-Mobility models. We incorporated natural lighting using Nvidia's ray tracing capabilities and simulated the Intel RealSense D435 as the RGB-D sensor within the environment, aiming to mimic real-world conditions. Each object was imported into our Isaac Sim environment, with one rotational joint axis, typically the primary axis, and its associated movable parts selected.

To achieve comprehensive sensor coverage, we implemented a systematic placement strategy.
Specifically, 
we uniformly distributed 128 RGB-D sensors over a sphere, capturing a wide range of viewing angles, with the object positioned at the sphere's center.
The camera poses were oriented to align with the central orthogonal coordinate system of the object.
Each camera generated an RGB image, a depth map, and a corresponding object mask. 
Subsequently, the joint axis underwent a rotation of either 90 degrees or to its maximum joint limit, 
transitioning the object to its post-interaction state. The sensor data capture process was then reiterated as described above.

The {\mma} dataset emerges as a comprehensive multi-view, multi-modal, and multi-state resource for articulated objects, laying a robust groundwork for training and evaluation in multi-modal interactive object reconstruction tasks. 
Additionally, its design allows for easy conversion into specialized single-object or single-modal datasets, thereby expanding its utility and reinforcing its relevance for various tasks within this domain.
\label{datasets}

\section{Experiments}

\subsection{Evaluation Metrics} 

\subsubsection{Geometric Reconstruction}  
To assess the quality of the reconstructed mesh for articulated objects, we employ the Chamfer Distance~(CD) metric. 
For evaluation purposes, all baseline methods are uniformly sampled to 4096 points. 
We also compute the CD metric exclusively for the point cloud associated with the movable component, 
providing insight into the method's precision in delineating movable parts.

\subsubsection{Joint Parameter}  Consistent with the evaluation metrics used in Ditto~\cite{jiang2022ditto}, we gauge the precision of the predicted rotational joint through a dual-metric approach. 
Initially, we determine the Angular Deviation~(Angle Error), which captures the variance in direction between the forecasted axis direction and the actual ground truth. 
Subsequently, we quantify the Axis Displacement~(Position Error), a metric that pinpoints the minimal spatial separation between the anticipated rotation axis and the ground truth, while also accounting for the pivot point's specific position.

\subsection{Baseline}
We introduce Ditto~\cite{jiang2022ditto} and propose two multi-view image-based methods for comparative analysis.
\subsubsection{Ditto}
Ditto~\cite{jiang2022ditto} uses neural representations for modeling articulated objects based on single-view point cloud pre- and post-interaction. 
The model is trained with ground truth 3D structure and joint parameters. 
We used a pre-trained Ditto, further training it on more categories with an MMArt-transformed dataset and tested it across 13 categories using depth map-derived point clouds.
\subsubsection{Arti Nvdiff}
Arti Nvdiff, built upon Nvdiffrec~\cite{munkberg2022extracting}, is a two-phase end-to-end network designed to identify joints in articulated objects. 
Initially, multi-view RGB images aid in reconstructing the object through grid deformation.
Then, the movable part undergoes a rigid rotational transformation using the corresponding joint. 
The joint parameters and the movable part are then refined by comparing the grid's rendering with post-interaction multi-view images. 
\subsubsection{Joint NeRF}  
Joint NeRF is rooted in NeRF-based reconstruction~\cite{mildenhall2021nerf} of objects both pre- and post-interaction. We employ the Marching Cube to derive meshes from radiance fields. Movable components are segmented as delineated in Section~\ref{sec:3.3_priori-seg}. Leveraging the keypoint algorithm~\cite{li2019usip}, we identify  candidate joints. The most fitting candidate, ascertained by the least RGB rendering discrepancy against the ground truth, dictates the ultimate joint parameter.
\label{benchmark}

\subsection{Main Results}
\begin{table*}[]
\centering
\resizebox{1\textwidth}{!}{
\begin{tabular}{c|c|ccccccccccccccc}
\toprule
& Method & Box & Laptop & Door & Safe & Microwave & Dishwasher & Storage & Eyeglasses & Staples & Washer & Oven & Fridge & Toilet & Mean \\
\midrule
\multirow{4}{*}{\makecell[cc]{Angle\\Error}} & Ditto & 3.425 & 3.463 & 1.975 & 2.551 & 2.517 & 4.011 & 4.031 & 3.072 & 3.115 & 1.412 & 4.514 & 5.342 & 2.776 & 2.947 \\
& Arti Nvdiff& 27.084 & 19.984 & 43.692 & 24.754 & 32.572 &  21.120 & 16.203 & 40.912 & 35.192 & 64.348 & 27.890 & 32.309 & 53.427 & 32.253 \\
& Joint NeRF & 7.245 & 0.801 & $\mathbf{0.795}$ & 13.580 & 11.984 & 9.039 & 19.253 & 82.298 & 49.945 & 59.434 & 6.023 & 40.561& 62.124 & 27.929 \\
& Ours & $\mathbf{0.922}$ & $\mathbf{0.784}$ & $1.001$ & $\mathbf{1.419}$ & $\mathbf{1.218}$ & $\mathbf{1.272}$ & $\mathbf{1.357}$ & $\mathbf{1.905}$ & $\mathbf{1.033}$ & $\mathbf{1.025}$ & $\mathbf{0.923}$ & $\mathbf{0.816}$ & $\mathbf{1.192}$ & $\mathbf{1.059}$ \\
\cmidrule(lr){1-16}
\multirow{4}{*}{\makecell[cc]{Position\\Error}} & Ditto & 0.203 & $\mathbf{0.050}$ & 0.177 & 0.201 & 0.179 & 0.276 & 0.186 & $\mathbf{0.073}$ & 0.161 & 0.224 & 0.173 & 0.291 & 0.230 & 0.158 \\
& Arti Nvdiff & 3.328 & 2.547 & 2.082 & 3.478 & 7.952 & 5.323 & 5.391 & 6.457 & 2.310 & 3.418 & 4.599 & 2.974 & 4.723 & 4.314 \\
& Joint NeRF & 0.337 & 0.084 &   $\mathbf{0.010}$& 0.493 & 0.512 & 0.329 & 0.957 & 3.429 & 2.605 & 4.253 & 3.507 & 3.185 &  6.632 & 2.025 \\
& Ours & $\mathbf{0.021}$ & 0.061 & $0.069$ & $\mathbf{0.149}$ & $\mathbf{0.134}$ & $\mathbf{0.202}$ & $\mathbf{0.116}$ & 0.124 & $\mathbf{0.125}$ & $\mathbf{0.187}$ & $\mathbf{0.109}$ & $\mathbf{0.152}$ & $\mathbf{0.048}$ & $\mathbf{0.101}$ \\
\bottomrule
\end{tabular}
}
\caption{Evaluations on the absolute errors of the articulation parameter estimation.}
\label{tab:quantit-joint}
\end{table*}

\begin{table*}[]
\centering
\resizebox{1\textwidth}{!}{
\begin{tabular}{c|c|ccccccccccccccc}
\toprule
& Method & Box & Laptop & Door & Safe & Microwave & Dishwasher & Storage & Eyeglasses & Staples & Washer & Oven & Fridge & Toilet & Mean \\
\midrule
\multirow{3}{*}{ \makecell[cc]{Whole\\Models}} & Ditto & 0.230 & 0.206 & 0.194 & 0.181 & 0.158 & 0.512 & 0.273 & 0.164 & 0.141 & 0.747 & 0.221 & 0.207 & 0.316 & 0.304 \\
& Joint NeRF & 0.142 & $\mathbf{0.085}$ & 0.088 & 0.109  & $\mathbf{0.105}$ & 0.113 & 0.112 & 0.107 & 0.097 & 0.100 & 0.095 & 0.106 & 0.120 & 0.106 \\
& Ours & $\mathbf{0.091}$ & 0.110 & $\mathbf{0.089}$ & $\mathbf{0.087}$ & $0.120$ & $\mathbf{0.081}$ & $\mathbf{0.099}$ & $\mathbf{0.103}$ & $\mathbf{0.088}$ & $\mathbf{0.079}$ & $\mathbf{0.090}$ & $\mathbf{0.074}$ & $\mathbf{0.097}$ & $\mathbf{0.086}$ \\
\cmidrule(lr){1-16}
\multirow{4}{*}{\makecell[cc]{Movable\\Parts}} & Ditto & 0.159 & 0.121 & 0.075 & 0.324 & 0.057 & 1.547 & 0.355 & 0.242 & 0.093 & 6.633 & 0.489 & 0.166 & 0.603 & 0.755 \\
& Arti Nvdiff & 1.502 & 0.993 & 1.663 & 0.938 & 1.179 & 0.984 & 0.846 & 1.114 & 1.657 & 0.858 & 1.340 & 1.226 & 1.478  &  1.056\\
& Joint NeRF & 0.865 & 0.522 & 0.508 & 0.743 & 0.512 & 0.962 & 1.136 & 0.968 & 1.044 & 0.624 & 0.793 & 1.114 & 0.988 & 0.787 \\
& Ours & $\mathbf{0.035}$ & $\mathbf{0.038}$ & $\mathbf{0.034}$ & $\mathbf{0.031}$ & $\mathbf{0.022}$ & $\mathbf{0.029}$ & $\mathbf{0.034}$ & $\mathbf{0.026}$ & $\mathbf{0.027}$ & $\mathbf{0.029}$ & $\mathbf{0.030}$ & $\mathbf{0.043}$ & $\mathbf{0.046}$ & $\mathbf{0.033}$ \\
\bottomrule
\end{tabular}
}
\caption{Evaluations on the Chamfer Distance of the 3D reconstruction of articulated objects.}\label{tab:quantit-recon}
\end{table*}

\subsubsection{Joint Parameter}  
From the quantitative results in Table~\ref{tab:quantit-joint}, our method outperforms other approaches in both numerical accuracy and stability. 
Arti Nvdiff is at a clear disadvantage compared to other methods, indicating that end-to-end implicit articulated object modeling still faces convergence challenges during training. 
The advantages of the Joint NeRF in some categories highlight the immense benefit of utilizing the geometric structure from pre- and post-interaction for object motion structure modeling. 
However, its inability to achieve joint optimization makes it overly reliant on the accuracy of keypoint detection, leading to failures on objects with complex geometric structures.
The visualization results in Fig.~\ref{fig:visual} reveal that, due to intra-class morphological variations, Ditto~\cite{jiang2022ditto} still misplaces the joint direction for some objects. 

\subsubsection{Geometric Reconstruction}  
Table~\ref{tab:quantit-recon} shows that our overall geometric reconstruction of the object is significantly better than Ditto's results and is close to the Joint NeRF method, which is based on NeRF~\cite{noguchi2022watch}.  
 The end-to-end Arti Nvdiff still struggles, while the segmentation results of Joint NeRF entirely depend on its joint estimation accuracy.
Although Ditto~\cite{jiang2022ditto}  has more geometric information due to its point cloud input compared to other methods, its single-view limitation and lack of texture hinder its generalization capability across all objects.
For movable part segmentation, our method has a distinct advantage over others, with a chamfer distance improvement of up to 96\% compared to the second-best, Ditto~\cite{jiang2022ditto}.
This is attributed to our joint optimization of the movable part and joints based on rendered RGB and the patch optimization of the image loss, which significantly enhances segmentation accuracy. 
Fig.~\ref{fig:visual} also reveals that our segmentation has almost no extraneous noise points.

\subsection{Ablation Study}

\begin{table}[]
\centering
\resizebox{0.95\columnwidth}{!}
{
 \begin{tabular}{c|c|cc|c}
        \toprule
        Case \# & Component & Ang. Err. & Pos. Err. & CD-Mov\\
        \midrule
       1 & Baseline & 32.253 & 4.314 & 1.056 \\
       2 & \#1 + Priori-Segmentation & 13.157 &  1.505 & 0.912 \\
       3 & \#1 + Candidate Joints & 2.147  & 0.172 &  0.219\\
       4 & \#2 + Candidate Joints & 1.302  & 0.163 &  0.116 \\
        \cmidrule(lr){1-5}
       5 & \#4 + PatchImage & 1.059  & 0.101 & 0.033\\
        \bottomrule
   \end{tabular}
}
\caption{Ablation studies results of different modules.}
\label{tab:abla}
\end{table}

Table~\ref{tab:abla} demonstrates the effectiveness of each component in the proposed {\ummm} framework. 
Case \#1 corresponds to the structure of Arti Nvdiff. 
Cases \#2 and \#3 highlight the significant role of Priori-Segmentation and Candidate Joints, derived from the geometric differences between pre- and post-interaction objects, in estimating the object's motion structure.
 Compared to the former, Candidate Joints play a more pronounced role due to their effective constraint on the 6-DoF joint position and direction in Case \#4. 
 Case \#5 shows that the PatchImage Loss further enhances the method's performance, especially for movable part segmentation, by computing a more detailed image loss.

\subsection{Real-world  Results}
\begin{figure}[t]
  \centering
  \includegraphics[width=0.9\linewidth]{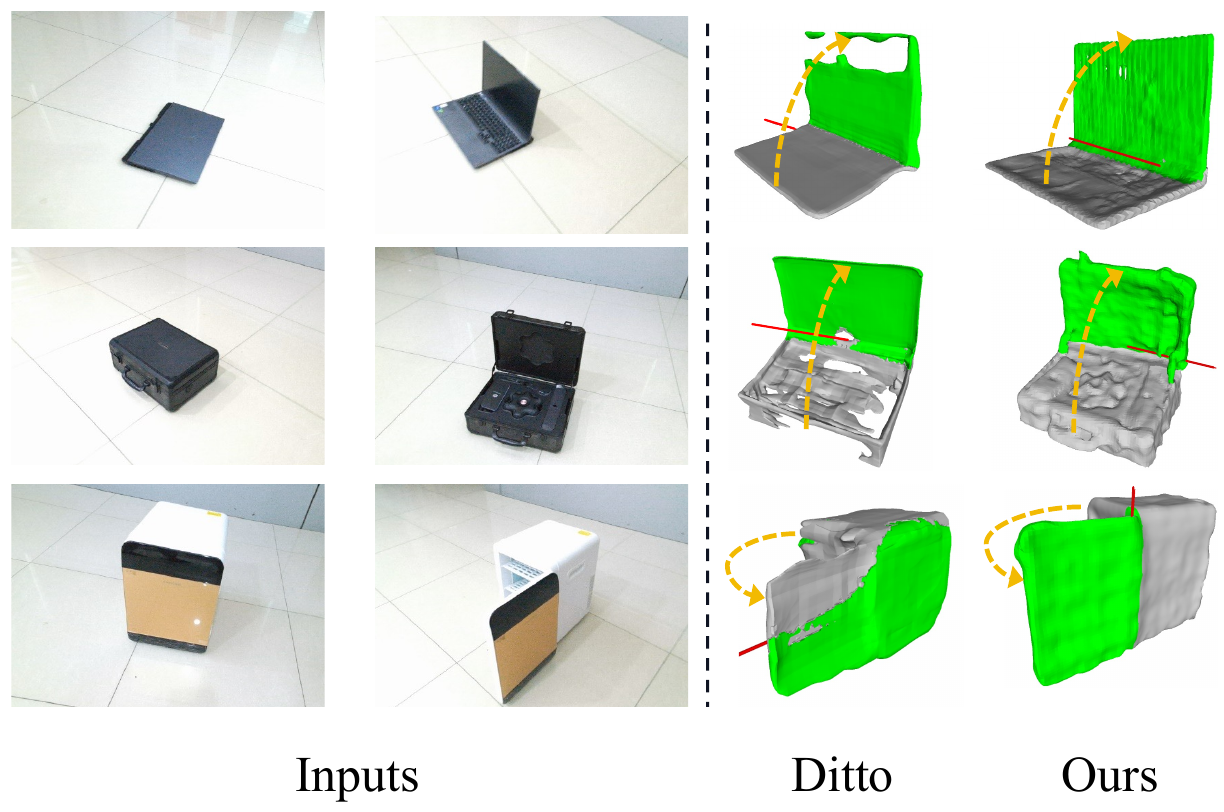}
  \caption{
Visualization results for real-world articulated objects.
  }
  \label{fig:realword}
\end{figure}

To validate the universality of our method, 
we tested it in real-world scenarios. 
Avoiding any potential for additional information, we solely used an iPhone 14 to capture videos centered on the object and extracted 64 images from different viewpoints. 
Subsequently, using these images and the corresponding poses from Colmap, we applied our proposed {\ummm} for modeling. As seen in Fig.~\ref{fig:realword}, our method consistently delivers outstanding performance across all objects.

\section{Conclusion}
We introduce the {\ummm} framework, 
a pioneering solution that, given only multi-view images captured before and after object interaction, 
achieves detailed textured 3D reconstruction and motion structure analysis of articulated objects without any labels or 3D models. 
This encompasses the segmentation of movable components and the estimation of joint parameters. 
Our approach outperforms supervised methods across all categories and objects. 
Furthermore, we present the {\mma} dataset, tailored for training and testing tasks related to articulated objects.

\clearpage
\bibliographystyle{IEEEtran}
\bibliography{reference}

\end{document}